\documentclass{isprs} 
\usepackage{subfigure}
\usepackage{setspace}
\usepackage{geometry}
\usepackage{epstopdf}
\usepackage[labelsep=period]{caption}
\usepackage[british]{babel} 
\usepackage[hang]{footmisc}

\usepackage{hyperref}
\usepackage{url}
\usepackage{booktabs}
\usepackage{amsfonts}
\usepackage{nicefrac}
\usepackage{microtype}
\usepackage{lipsum}
\usepackage{graphicx}
\usepackage{natbib}
\usepackage{doi}

\usepackage{amsmath}
\usepackage{amsfonts}
\usepackage{amssymb}
\usepackage{array}
\usepackage{multirow}
\usepackage{xcolor}
\usepackage{soul}
\usepackage{svg}
\usepackage{adjustbox}
\usepackage{tabularray}
\usepackage{fix-cm}

\usepackage{courier}

\usepackage{xspace}
\makeatletter
\DeclareRobustCommand\onedot{\futurelet\@let@token\@onedot}
\def\@onedot{\ifx\@let@token.\else.\null\fi\xspace}

\makeatother

\begin{document}



\title{GSR4B: Biomass Map Super-Resolution with Sentinel-1/2 Guidance}

\date{}


\author{
Kaan Karaman\textsuperscript{1}\thanks{Corresponding author}~, 
Yuchang Jiang\textsuperscript{1}, 
Damien Robert\textsuperscript{1}, 
Vivien Sainte Fare Garnot\textsuperscript{1}, 
Maria João Santos\textsuperscript{2}, 
Jan Dirk Wegner\textsuperscript{1}}

\address{
\textsuperscript{1 }EcoVision Lab, DM3L, University of Zurich, Switzerland\\
{\tt\small \{kaan.karaman, yuchang.jiang, damien.robert, jandirk.wegner\}@uzh.ch, vsaint@ics.uzh.ch}\\
\textsuperscript{2 }Department of Geography, University of Zurich, Switzerland\\
{\tt\small maria.j.santos@geo.uzh.ch}\\
}



\keywords{GSW 2025, Biomass Estimation, Guided Super-Resolution}
\maketitle


\begin{abstract}

Accurate Above-Ground Biomass (AGB) mapping at both large scale and high spatio-temporal resolution is essential for applications ranging from climate modeling to biodiversity assessment, and sustainable supply chain monitoring. 
At present, fine-grained AGB mapping relies on costly airborne laser scanning acquisition campaigns usually limited to regional scales. 
Initiatives such as the ESA CCI map attempt to generate global biomass products from diverse spaceborne sensors but at a coarser resolution. 
To enable global, high-resolution (HR) mapping, several works propose to regress AGB from HR satellite observations such as ESA Sentinel-1/2 images. 
We propose a novel way to address HR AGB estimation, by leveraging both HR satellite observations and existing low-resolution (LR) biomass products. 
We cast this problem as Guided Super-Resolution (GSR), aiming at upsampling LR biomass maps (sources) from $100$ to $10$ m resolution, using auxiliary HR co-registered satellite images (guides).
We compare super-resolving AGB maps with and without guidance, against direct regression from satellite images, on the public BioMassters dataset.
We observe that Multi-Scale Guidance (MSG) outperforms direct regression both for regression ($-780$ t/ha RMSE) and perception ($+2.0$ dB PSNR) metrics, and better captures high-biomass values, without significant computational overhead.
Interestingly, unlike the RGB+Depth setting they were originally designed for, our best-performing AGB GSR approaches are those that most preserve the guide image texture. 
Our results make a strong case for adopting the GSR framework for accurate HR biomass mapping at scale.
Our code and model weights are made publicly available \footnote{\url{https://github.com/kaankaramanofficial/GSR4B}}.

\end{abstract}

The increasing effects of climate change in recent decades highlight the urgency to study ecological and earth system processes, as well as to develop methods that provide a robust and systematic quantification of ecological parameters \citep{Baidya2024BibliometricClimateChange, Tuia2021Agenda, Reichstein2019DataDrivenEarth, Irrgang2021AI_in_EarthSystem}.
In particular, Above Ground Biomass (AGB) is both an essential climate variable \citep{BiomassBook2009, Canadell2008ForestIsImportant} for carbon flux modeling, and an essential biodiversity variable \citep{skidmore2021priority} for ecosystem modeling. 
Despite recent advancements in sensor technology, it remains challenging to accurately and cost-effectively measure AGB over large areas and with high spatial resolution. 
AGB mapping efforts are typically faced with a trade-off between spatial extent and resolution \citep{BioMassters23}.
High-resolution (HR) AGB maps are typically obtained from statistical allometric equations \citep{Smith1983AllometricEquation} estimating the biomass from other in-situ measurements of vegetation characteristics \citep{brede2022non} (e.g., canopy height, diameter at breast height).
While this approach can yield HR AGB maps at metric resolution for areas in the order hectares, its prohibitive operational costs prevent scaling to continental or global extents. 
Meanwhile, AGB mapping projects such as the European Space Agency (ESA) Biomass Climate Change Initiative (CCI) \citep{ESA_CCI} produce maps with global coverage, but with a relatively low spatial resolution of $100 \times 100$ meters. 
Other approaches choose to directly estimate AGB from globally-available HR satellite imagery \citep{li2020forest, lang2021high, kanmegne2022estimation}, yielding predictions of up to $10 \times 10$ meters resolution but tend to underestimate the content of high biomass zones. 

\begin{figure}[hbt!]
    \centering
    \includegraphics[width=\linewidth]{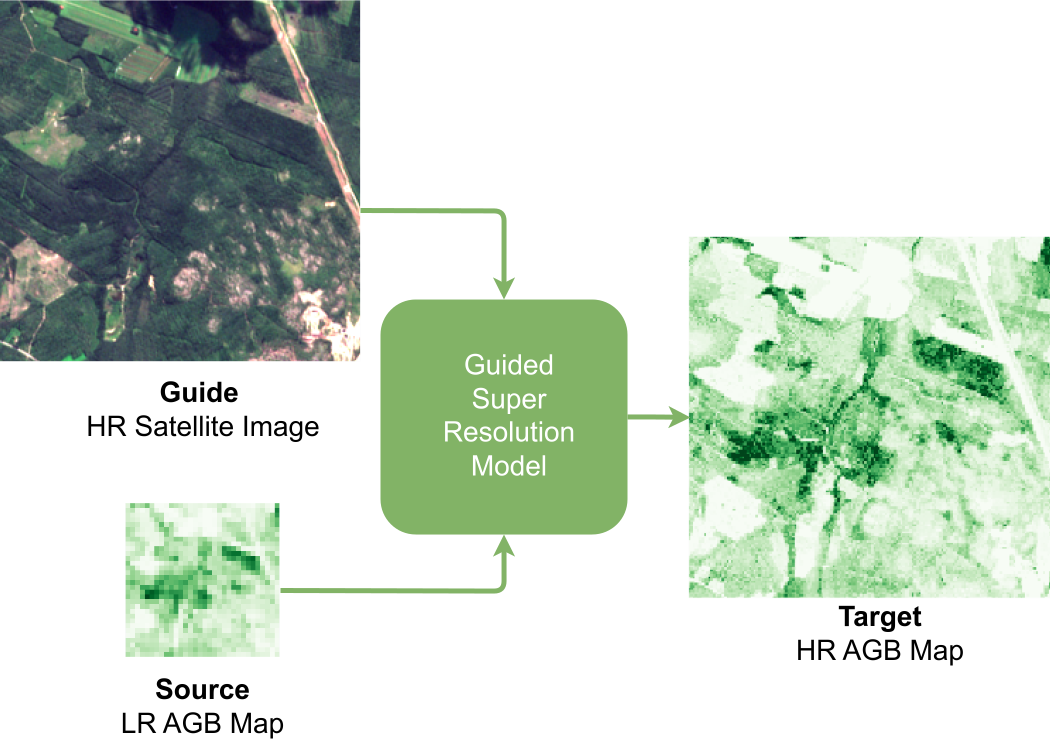}
    \caption{
    Given a LR AGB map and co-registered HR satellite imagery,
    our method predicts the upsampled HR biomass map.
    Increasing AGB values are represented from white to green.
    }
    \label{fig:Teaser}
\end{figure}

Unlike previous studies, we take as input low-resolution (LR) and HR data together, allowing our model to extract and fuse low- and high-frequency (LF \& HF) information for more accurate HR outputs alongside less underestimation at high biomass zones.
We leverage both low-resolution AGB products (could be global extent) and HR satellite observations (from Sentinel-1/2) to estimate HR AGB maps ($10 \times 10$ meter resolution), as illustrated in Figure \ref{fig:Teaser} where the scale factor is $8$.
This task is referred to as Guided Super-Resolution (GSR) and has been extensively explored in the computer vision literature, especially in the context of RGB $+$ Depth datasets, where a depth map is upsampled using guidance from an RGB image \citep{Zhong2023GuidedSuperResSurvey}. 
Here, we draw on this literature and explore the performance of established GSR approaches in a novel application setting: AGB map super-resolution with satellite imagery guidance.
We conduct our experiments on the BioMassters dataset \citep{BioMassters23} and extensively compare four different GSR methods, simple upsampling with conventional interpolation techniques, and direct biomass regression from satellite imagery.
Our results show that casting the problem as a GSR task yields superior performance. 
The contributions of this study can be summarized as follows:
\begin{enumerate}
    \item 
        \textit{Introducing the problem of GSR for biomass (GSR4B).} 
        We propose an original way to address AGB mapping, which to the best of our knowledge has not been previously explored. 
    \item 
        \textit{Benchmarking GSR approaches.}
        We compare the performance of both learned and non-learned GSR techniques to AGB map upsampling on a public dataset, with reproducible experiments.
        We find that deep learning (DL) based methods with texture-copying produce the best results and outperform direct regression both in terms of regression ($-780~\text{t}/\text{ha}$ RMSE, $-570~\text{t}/\text{ha}$ MAE) and perception ($+2.0$ dB PSNR, $+0.07$ SSIM) metrics. In particular, the GSR models reduce the error at high AGB values.
\end{enumerate}

\subsection{Biomass Estimation}
\label{ssec:Related:BiomassEstimation}
Despite its critical role in understanding Earth system processes that inform environmental policy, biomass data remains challenging to obtain both accurately and at large scale. 
Recent advancements focus on two key areas: data acquisition techniques and model design for estimating AGB from remote sensing data. 

\paragraph{Biomass datasets.} 
Early methods for estimating biomass have relied on destructive in-situ measurements, where field teams physically collected data from specific areas to obtain AGB \citep{Baskerville1972AllometricEquation}. 
Despite being the only accurate way to assess aboveground biomass, this approach can only be limited to a few trees at once located in accessible areas \citep{Kumar2017RemoteSensingABG}. 
Airborne sensors \citep{Tian2023ReviewForRM_AGB} circumvent these limitations by indirectly estimating biomass through other tree parameters such as canopy height, which are then combined with allometric equations tailored to specific forest types \citep{Smith1983AllometricEquation, Ketterings2001AllometricEquation}. 
This approach offers high resolution and precision but still remains impractical and costly for very large areas or global-scale applications.

As a result, recent AGB estimation works strive to leverage the global, geospatial information captured by spaceborne sensors.
However, the AGB maps produced from open satellite images such as Sentinel-1$/$2 tend to have lower resolution and accuracy than maps based on airborne and in-situ measurements. 
To combine the strengths of the data acquired by different sensors, some recent datasets \citep{BioMassters23, Ghjulia2024AGBD, Liu2023SouthWestChinaAGBDataset, Leonhardt2022ProbBiomassCGAN} have been developed to train data-driven methods that improve satellite-data-derived AGB estimates.
\citet{BioMassters23} focuses on a single type of forest in a localized region of Finland, providing detailed HR AGB maps. 
In contrast, \citet{Ghjulia2024AGBD} samples AGB data more sparsely from diverse locations globally. 
Both datasets co-register satellite data with AGB maps derived from airborne sensors, allowing models to process the satellite images to predict HR AGB maps.

Furthermore, thanks to the recent advances in spaceborne sensor technology, the ESA launched the CCI \citep{ESA_CCI} to generate global-scale datasets for climate prediction, including AGB data. 
Unlike localized datasets like \citet{BioMassters23} or sparsely sampled ones like \citet{Ghjulia2024AGBD}, the ESA CCI project biomass product has global coverage, including remote and under-monitored forests. 
This data product is particularly valuable for studying areas that typically lack regular forest monitoring. 
The ESA CCI biomass data is made available at higher temporal sampling, enabling the tracking of changes in biomass and their impact on biodiversity.

In comparison to other spaceborne biomass data, the ESA CCI biomass data produces AGB maps at a spatial resolution of $100 \times 100$ meters, one order of magnitude lower than the $10 \times 10$ meter resolution available in other datasets, and that of Sentinel-1 \& 2 satellite data. 
As such, to date, AGB datasets face a trade-off between spatial resolution, high accuracy, and high temporal resolution. 
Here, we demonstrate the advantages of combining low-resolution AGB data with higher-resolution satellite imagery, to produce AGB maps of both higher accuracy and resolution than existing methods. 

\paragraph{Biomass estimation from remote sensing data.}
Early approaches to biomass estimation from remote sensing data rely on non-parametric models such as \citet{Beaudoin2014MODIS_KNN} and kernel-based methods, as discussed by \citet{Tian2023ReviewForRM_AGB}. 
More recently, deep learning became the prominent approach, similar to the trends in other areas \citep{Tian2023ReviewForRM_AGB}. 
Several published DL-based models reach a high performance in estimating biomass \citep{Lu2016RemoteSensingABG_Survey, Weber2024RS_ABG_CH_Estimation} or related ecological parameters such as canopy height \citep{Nico2023CanopyHeightNatureEcology, Yuchang2023CanopyHeight, Astola2021CanopyHeight}.

Most recent DL-based models for biomass have been developed for localized regions, primarily due to the limited availability of training labels, and the availability of Sentinel-2 given with its HR and free public access. 
For example, \citet{Yuchang2023CanopyHeight} uses Sentinel-2 data to generate canopy height maps at a $10 \times 10$ meter spatial resolution. 
\citet{puliti2020BiomasS2DEM} combines Sentinel-2 and Digital Elevation Model (DEM) data to model biomass in Norway, while \citet{Alex2023ResNextCountryWide} integrates Sentinel-1 \& 2 data to estimate forest structure also in Norway.
\citet{Liu2023SouthWestChinaAGBDataset} also uses a combination of Sentinel-1 \& 2 data to estimate biomass in China. 
Other approaches combine Sentinel-2 with L-band radar observations to improve their estimation \citep{vafaei2018improving}.
Despite the HR biomass maps achieved using these approaches, they often apply direct regression between satellite imagery and target parameters, overlooking the potential of existing auxiliary biomass products. 
For instance, the ESA CCI biomass product, with its global coverage at a $100 \times 100$ meter resolution, offers significant opportunities for further exploration and integration, yet remains underutilized in favor of higher-resolution images for direct regression.

In this work, benchmark AGB estimation methods based on established models that demonstrated significant success in biomass estimation or related ecological parameters directly from satellite imagery. 
Two DL-based architectures stand out: U-Net \citep{Ronneberger2015UNet} and ResNeXt \citep{Xie2017ResNext}. 
More specifically, we evaluate variants of those architectures developed for related remote sensing tasks. 
Namely, we focus on the U-TAE \citep{Vivien2021UTAE}, which is the strongest baseline on the BioMassters white paper, and the ResNeXt model of \citet{Alex2023ResNextCountryWide}, which is optimized for canopy height regression.

\subsection{Guided Super-Resolution}

In this paper, we propose an alternative approach for estimating biomass also leveraging LR maps. 
We reformulate the problem as a GSR problem. 
This involves merging satellite images with LR biomass maps to maximize the available information for a specific region. 
Technically, this problem is analogous to guided depth super-resolution (GDSR) \citep{Zhong2023GuidedSuperResSurvey}. 
For the sake of analogy, we replace natural images with satellite images and LR depth maps with LR biomass maps, while keeping the input-output structure and model objectives the same. 
In the next section, we review the rationale we used to justify this approach.

\paragraph{GDSR approaches.}
Starting with one of the simplest GDSR methods, Joint Bilateral Upsampling (JBU) \citep{Kopf2007JBU}
is the first to utilize a bilateral filter to restore an HR image by combining information from both the HR guide image and LR source image. 
Several subsequent studies \citep{Liu2013JGU, Jevnisek2017CoOccurance} propose to extend this method.
According to \cite{Zhong2023GuidedSuperResSurvey}, early versions of guided filters show better performance than bilateral approaches.
They however have another limitation associated with assuming a linear relationship between the gradients of the guide image and their HR outputs, which causes texture-copying, that is, transferring irrelevant textures from the guide to the prediction. 
Indeed, in RGB$+$Depth datasets, depth masks typically have a much smoother texture than the guide RGB images. 
Subsequently, the very first and simplest DL-based approach is developed, namely Deep Multi-Scale Guidance (MSG) \citep{Hui2016MSG}. 
This architecture utilizes deconvolutional blocks to upsample the source map while merging features from the guide image at different resolution levels. 
Over time, more advanced DL models are proposed to increase the capacity to mitigate the texture-copying issue. 
For example, Pixel-to-Pixel Transform (P2P) \citep{Lutio2019P2P} introduces a shallow DL model and defines an unsupervised loss function to train it sample by sample. 
Fast Depth Map Super-Resolution (FDSR) \citep{He2021FDSR} focuses on the frequency components of the information in the guide and source images. 

\paragraph{Super-resolution methods.} GSR is an extension of the broader task of image super-resolution. 
The most common and basic methods in computer vision literature for this purpose are nearest-neighbor, bilinear, and bicubic interpolations.
Next, DL-based approaches gain prominence in this field as well. 
They aim to train a model that integrates domain-specific prior knowledge with local and global statistics from LR input maps to produce HR, photo-realistic outputs \citep{Li2024DLSingleSRSurvey}. 
Early research primarily explores CNN-based architectures. 
However, more recent studies are heavily influenced by transformer-based architectures, leveraging their ability to capture long-range dependencies for improved performance. 
SwinIR \citep{Liang2021SwinIR} achieves notable success by incorporating Residual Swin Transformer blocks, and subsequent models like SwinFIR and HAT have refined this approach. 
SwinFIR \citep{Zhang2022SwinFIR} revisits the limitations of SwinIR, improving the performance by Fast Fourier Convolution, while HAT \cite{Chen2023HAT} enhances performance by processing a larger pixel context to enhance image quality further.

\subsection*{Problem Formulation}

The goal of GSR4B is to estimate HR AGB maps by leveraging LR source AGB maps alongside HR satellite images as depicted in Figure \ref{fig:Teaser}. 
Intuitively, we aim to combine the HF components of the HR satellite inputs with the LF information from the LR sources so as to reconstruct the HR version of the source maps.
Our problem can be expressed by Equation \eqref{eq:MainAimEquationForL1Norm}.

\begin{equation}
    \label{eq:MainAimEquationForL1Norm}
    \theta^* = \arg \min_\theta \sum_{i=0}^{N-1} ||\xi_\theta(S_i, G_i) - Y_i||_1~,
\end{equation}

where $\xi_\theta(\cdot, \cdot): \mathbb{R}^{H/\alpha \times W/\alpha} \times \mathbb{R}^{H \times W} \rightarrow \mathbb{R}^{H \times W}$ is the model that predicts the HR map with the help of the guide image and the source map.
$N$ and $\theta$ denote the number of test samples, and the trainable or fixed parameters of the model, respectively. 
The other notations are shown in Table \ref{tab:Notations} for clarifying the mathematical formulations in the following sections.

\begin{table}[hbt!]
    \centering
    \begin{tabular}{cc}
        \toprule
        Notation & Meaning \\
        \cmidrule{0-1}
        $H, W \in \mathbb{Z}^+$ & Height \& width of HR \\
        $\alpha \in \mathbb{Z}^+$ & Scaling factor \\
        $S \in \mathbb{R}^{H/\alpha \times W/\alpha} $ & Source map in LR \\
        $G \in \mathbb{R}^{H \times W} $ & Guide image in HR \\
        $Y \in \mathbb{R}^{H \times W}$ & Target map in HR \\
        $\hat{Y} \in \mathbb{R}^{H \times W}$ & Predicted map in HR \\
        \bottomrule
    \end{tabular}
    \caption{Notations used throughout this paper.}
    \label{tab:Notations}
\end{table}

\subsection*{Guided Super-Resolution Methods}
We investigate four GSR-based approaches: JBU, MSG, P2P, and FDSR.

\paragraph{JBU.}
\citet{Kopf2007JBU}, a non-learning-based GSR method, utilizes a bilateral filter to restore an HR map by combining information from both the HR guide image and the LR source map. Its formula is presented in Equation \eqref{eq:JBU}.

\begin{equation}
\label{eq:JBU}
    \hat{Y}_p = \frac{1}{k_p} 
    \sum_{q_\downarrow \in N_p} 
    {S_{q_\downarrow} \cdot f( || p_\downarrow - q_\downarrow||_2 ) \cdot g( || G_p - G_q||_2)}~, 
\end{equation}

where  $p$ and $q$ denote the pixel locations in the HR grid. $k_p, N_p, f(\cdot)$ and $g(\cdot)$ represent the normalization value around pixel $p$, neighborhood (support) pixels, spatial and range kernels, respectively. In the original paper \citep{Kopf2007JBU}, both kernels are chosen as truncated Gaussian PDFs, and the distance function is defined using the $\ell_2$-norm. Although it is not a data-driven approach, the model still requires careful tuning of hyperparameters, such as the standard deviations for the Gaussian PDFs used in range and spatial kernels.

\paragraph{MSG.} 
In this model \citep{Hui2016MSG}, the guide and source images are passed through convolutional and deconvolutional blocks \citep{Zeiler2010Deconvolutional}, respectively. These blocks are arranged in a reverse manner, allowing the feature maps of the same resolution to be concatenated in fusion blocks. Specifically, the output of the first convolutional layer is merged with the output of the last deconvolutional block. Finally, the model includes a reconstruction part, which predicts the upsampled version of the source image. Because MSG is introduced earlier, its ability to handle texture-copying remains limited among the other more sophisticated DL-based GDSR methods below.

\paragraph{P2P.}
The key contribution of \citet{Lutio2019P2P} is defining an unsupervised loss function, as shown in Equation \eqref{eq:P2PUnsupervisedLoss}. 

\begin{equation}
    \label{eq:P2PUnsupervisedLoss}
    \mathcal{L} = \sum_{p \in D} ||S_p - <f_\theta(G, X)>_{\alpha, p} ||_2 + \lambda || \theta ||_2^2~,
\end{equation}

where $<\cdot>_{\alpha, p}$ is the downsampling operator when $\alpha, p$ are scaling factor and the pixel location. $\lambda$ denotes the regularization constant. Also, $X$ are the corresponding grid values of the pixel locations in the HR map. 
Like JBU, this technique does not follow the typical training paradigm; however, the DL-based model is trained during the inference phase. In this approach, the model aims to learn the relationship between the pixel-wise distributions of the guide image and the target HR map, as well as the pixel locations, by minimizing the loss values, calculated by using a single source-guide image pair. These loss values depend solely on the LR source image. Once the model overfits, the last predicted HR map is considered the final output. Due to its inference-time training procedure, this method is significantly slower compared to the other methods discussed in this section.

\paragraph{FDSR.}
The core idea behind \citet{He2021FDSR} is to utilize octave convolution blocks \citep{Chen2019OctaveConv} to separate information into different spatial frequencies within an image. By doing so, the method reduces channel redundancy and effectively merges the HF components of the guide images with the LR components of the source maps, ultimately reconstructing the HR target map. Then, FDSR further enhances the capability to tackle texture-copying with avoiding unwanted textures, which are filtered out by octave convolution.

\subsection*{Baselines}
We compare the selected GSR methods to two groups of baselines. 
The first group focuses on biomass estimation (BE) by directly regressing biomass from input satellite images. 
For this, we employ U-Net of \citet{Vivien2021UTAE} and ResNeXt of \citet{Alex2023ResNextCountryWide}, using only guided satellite images as inputs. 
The second group consists of super-resolution (SR) techniques which, unlike GSR, rely solely on LR maps to reconstruct HR outputs without any guidance.
The SR group includes three interpolation techniques, namely \textit{nearest-neighbor}, \textit{bilinear}, and \textit{bicubic}, as well as a customized MSG backbone deprived of guidance.

\subsection*{Dataset}
According to \citet{Ghjulia2024AGBD}, several datasets provide various benefits for research in biomass estimation. For this study, we opt for a publicly available dataset, containing AGB maps with co-registered satellite images. BioMassters \citep{BioMassters23} fits these criteria. 

\paragraph{Dataset description.}
The BioMassters dataset contains $13,000$ AGB map patches ($8,689$ of them are in the training set), of shape $256 \times 256$ pixels at a $10 \times 10$ meter resolution. The AGB values contained in those maps are derived from LIDAR sensing. These maps are co-registered with Sentinel-1 \& 2 images at the same resolution. The combined satellite observations are composed of 15 channels ($4$ from Sentinel-1, $10$ from Sentinel-2, and a cloud probability mask). Overall, the dataset covers roughly $8.5$ million hectares in Finland. Although the dataset provides $12$ time steps for each sample, we consider only the last ones to limit our problem as single image estimation or super-resolution, and leave multi-temporal guided biomass super-resolution for further research\footnote{As a result, the U-TAE architecture defaults to a U-Net model.}. 
Since the test set used in competition is not public, we randomly divide the publicly available training dataset into fixed training, validation, and test sets with ratios of $0.6$, $0.2$, and $0.2$, respectively.

\paragraph{Downsampling for source maps.}
BioMassters provides only HR satellite images and a single HR biomass map per sample. We however need to have LR versions of the biomass maps. Since the BioMassters dataset does not provide geolocation information, we cannot retrieve the corresponding ESA CCI AGB maps. Therefore, we downsample the target maps to create the source images, as typically done for single image super-resolution (SISR) \citep{Moser2023SRHitchhikerGuide, Li2024DLSingleSRSurvey}. This enables us to showcase a proof-of-concept of GSR4B.

We apply average pooling with an $\alpha \times \alpha$ kernel. 
For the choice of $\alpha$, we aim at replicating the GSR setup by using satellite images as guides and ESA CCI AGB maps as LR sources to predict HR AGB maps, as provided by BioMassters. While the ESA CCI AGB maps have a resolution of $100 \times 100$ meters, the BioMassters dataset uses a higher resolution of $10 \times 10$ meters. To align with the other GSR tasks, we select a scaling factor of $8$ rather than the ideal factor of $10$ since the power-of-2 scaling is commonly used in GSR approaches. 
All in all, this scaling factor allows us to downsample the ground truth maps from the size of $256 \times 256$ to the LR source map in the size of $32 \times 32$. 

\subsection*{Implementation Details}
\label{ssec:ImplementationDetails}
We conduct all experiments using the PyTorch library \citep{Ansel2024PyTorch} and an A100 GPU. While implementing the models, we adhere as closely as possible to the original implementations. However, because our guide images have $15$ channels (compared to $3$ channels for RGB images in the HR depth estimation task), we replace the initial convolutional layers' input channel from $3$ to $15$.
For models without available code, we implement them from scratch, ensuring they work similarly to the descriptions in the original papers. For more details, refer to these implementations in our GitHub repository.
We select the hyperparameters recommended in the original papers of the DL models and JBU. However, due to the large input size ($15 \times 256 \times 256$), we reduce the batch size of the ResNeXt model to $4$. All the implemented approaches are available in our GitHub repository\footnote{\url{https://github.com/kaankaramanofficial/GSR4B}}.

\subsection*{Evaluation Metrics}
\label{ssec:PerformanceMetrics}
We evaluate the performance of the different approaches using mean absolute error (MAE) and root mean square error (RMSE), commonly used in biomass estimation, as well as peak signal-to-noise ratio (PSNR), and structural similarity index measure (SSIM), typically used in super-resolution.

\begin{table}[t!]
    \centering
    \begin{tabular}{lccccc}
        \toprule
        & MAE$_\downarrow$ & RMSE$_\downarrow$ & PSNR$_\uparrow$ & SSIM$_\uparrow$ & TP$^\dagger_\uparrow$ \\
        & \textit{$\text{t}/\text{px}$} & \textit{$\text{t}/\text{px}$} & \textit{dB} & & \textit{Mpix$/$s} \\
        \cmidrule{2-6}
        \textit{GSR}$^\bigstar$ \\ 
        \cmidrule{1-1}
        MSG & \textbf{16.2} & \textbf{29.8} & \textbf{50.8} & \textbf{0.71} &  53.2 \\
        FDSR & 18.4 & 33.4 & 49.8 & 0.64 & 240.8 \\
        P2P & 26.6 & 46.3 & 47.0 & 0.55 & 0.3 \\
        JBU & 25.3 & 42.6 & 47.7 & 0.46 & 0.9 \\
        \\
        \textit{SR}$^\clubsuit$ \\ \cmidrule{1-1}
        MSG$^\text{ng}$ & 21.5 & 37.8 & 48.7 & 0.55 & 116.8 \\
        Nearest & 25.3 & 42.6 & 47.7 & 0.46 & \textbf{262 K} \\
        Bilinear & 25.3 & 41.1 & 48.0 & 0.46 & 82 K \\
        Bicubic & 24.0 & 39.5 & 48.4 & 0.49 & 37 K \\
        \\
        \textit{BE}$^\spadesuit$ \\ 
        \cmidrule{1-1}
        U-Net & 21.9 & 37.6 & 48.8 & 0.64 & 19.0 \\
        ResNeXt & 22.5 & 39.0 & 48.4 & 0.62 & 0.8 \\
        \bottomrule
    \end{tabular}
    \caption{
    Quantitative performance and throughput of AGB predictions from Guided Super-Resolution$^\bigstar$, Super-Resolution$^\clubsuit$, and direct Biomass Estimation$^\spadesuit$ techniques.
    We find that MSG outperforms all alternatives under all metrics. 
    \textbf{Best Performance}.
    $^\dagger$: Throughput for inference.
    $^\text{ng}$: No-guidance.
    px$~=100~\text{m}^2$.
    }
    \label{tab:Results}
\end{table}

\paragraph{Overall performance.} 
We present our experimental results in Table \ref{tab:Results}. We observe that the MSG approach outperforms all other methods across metrics. 
In terms of RMSE, MSG improves over direct biomass regression by $7.8~\text{t}/\text{px}$ ($=78~\text{kg}/\text{m}^2 = 780~\text{t}/\text{ha}$), which amounts to roughly a $20\%$ reduction of RMSE. 
This suggests, that addressing biomass estimation as a GSR problem can unlock significant gains in prediction accuracy compared to direct regression. 
MSG also improves by $9.7~\text{t}/\text{px}$ ($=970$ t/ha) RMSE compared to the best super-resolution approach (except MSG$^\text{ng}$, not taking any guide images) and by $12.8~\text{t}/\text{px}$ ($=1280$ t/ha) RMSE compared to JBU. 
While the interpolation baselines require less computational resources, these results show the benefit of training DL approaches for the task at hand. 
In the rightmost column, we provide the inference time measurements on a single A100 GPU. Despite these numbers being highly dependent on their implementations, we can still conclude that GSR methods do not need notably more time to process the additional LR source information. As a matter of fact, thanks to its light design, MSG is approximately three times faster than the U-Net baseline.

\begin{figure}[b!]
    \centering
    \includegraphics[width=\linewidth]{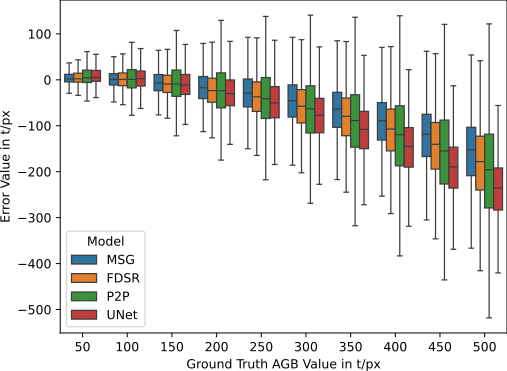}
    \caption{The box plot shows the errors of the predictions versus the ground truth AGB values (without outliers for simplicity). For each box, we randomly select $10~000$ pixels from the test set. px$~=100~\text{m}^2$, Error~$=$~Model~Prediction~$-$~Ground~Truth~AGB}
    \label{fig:ResidualPlot}
\end{figure}

\paragraph{Better performance at high AGB.} 
A widespread challenge in AGB estimation from remote sensing data is the underestimation of high biomass values \citep{BioMassters23, Ghjulia2024AGBD}.
We explore the distribution of estimation errors for several methods in Figure \ref{fig:ResidualPlot}.
We find that all GSR methods have a lower underestimation error than a U-Net directly regressing biomass from satellite imagery.
In particular, MSG estimates consistently show the lowest error across the entire biomass distribution range, with a variance comparable or lower than U-Net.
These results highlight the potential of GSR4B for accurately monitoring crucial high-AGB areas, which generally coincide with both high carbon stock, and high biodiversity.

\paragraph{Comparison between GSR methods.} 
Of the four tested GSR methods, MSG largely outperforms P2P and JBU by over $10~\text{t}/\text{px}$ ($=1000$ t/ha) RMSE.
MSG also performs better than FDSR by a margin of a $3.6~\text{t}/\text{px}$ ($=360$ t/ha) RMSE.
We find these results particularly interesting. Indeed, MSG is an earlier and simpler approach for GSR, and one could expect it to be outperformed by more elaborate approaches such as FDSR. 
As a matter of fact, when evaluated on the NYU-Depth V2 \citep{SilbermanECCV2012NYU_V2} dataset for RGB $+$ Depth data, FDSR achieves a performance almost twice better than MSG (see Table 1 in \citet{He2021FDSR}). 
This highlights an important discrepancy between GSR for biomass, and GSR for depth data. 
We suspect that earlier GSR approaches that do not mitigate the texture-copying issue are actually better suited for biomass, as they better preserve the HR details of the guide satellite image, which is useful for the HR biomass map. 

\paragraph{Importance of guiding.} 
As MSG consistently outperforms the other models, we use this model to evaluate the contribution of Sentinel-1 \& 2 guiding to the overall performance. 
We modify the MSG architecture and remove all convolutional blocks processing the HR satellite image and only give the source LR biomass map as input. 
In other words, we turn the MSG architecture into a simple SISR model, as explained in its original paper \citep{Hui2016MSG}.
Removing the HR guide image for the \textit{MSG no-guide} model (MSG$^\text{ng}$) results in a drop of $8.0~\text{t}/\text{px}$ ($=800$ t/ha) RMSE compared to \textit{MSG with guidance} (MSG).
This performance drop places MSG$^\text{ng}$ at a similar performance level as the direct biomass estimation baseline U-Net. This suggests that the source LR biomass map and the guide HR image play an equally important role in GSR4B. 

\begin{figure*}[htb!]
    \centering
    \includegraphics[width=0.8\textwidth]{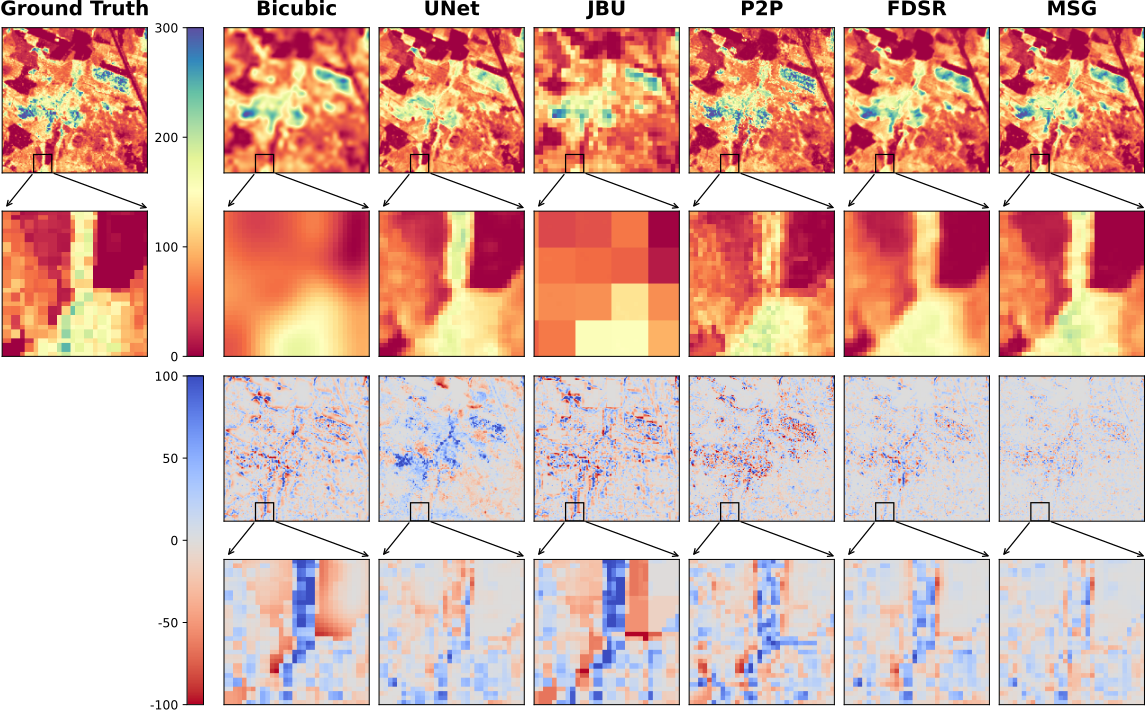}    
    \caption{The first two rows show the predicted AGB maps from various methods alongside the ground truth high-resolution maps of a random test sample. Lower to higher AGB values range from red to blue. The last two rows show the residuals between the reference and predicted maps. Joint Bilateral Upsampling (JBU), Pixel-to-Pixel Transform (P2P), Fast Depth Map Super-Resolution (FDSR) and Deep Multi-Scale Guidance (MSG) models are originally the guided depth super-resolution methods, tested on the biomass dataset.
    }
    \label{fig:PredictionsForFixedSamples}
\end{figure*}

\paragraph{Qualitative results.} 
In addition to the quantitative analysis, we also provide qualitative comparisons by selecting a random sample from the test set and plotting the outputs of different models used in the experiments. First, we compare the best models from each type of approach: direct biomass estimation, interpolation, and GSR. As shown in Figure \ref{fig:PredictionsForFixedSamples}, bicubic interpolation produces overly smooth maps compared to the other methods, while MSG captures finer textures (shown in cyan/green color) better than U-Net.

Again in Figure \ref{fig:PredictionsForFixedSamples}, we further compare different GSR approaches. Similar to its quantitative performance, JBU produces HR maps resembling those from nearest-neighbor interpolation. FDSR results in overly smooth outputs compared to P2P and MSG. P2P, in contrast, creates maps with more texture, but these textures do not align with the target map, indicating that P2P tends to introduce texture that is not related to the details contained in the ground truth map. MSG avoids this issue, retaining details that are closer to the ground truth HR map.

\begin{figure}[htb!]
    \centering
    \includegraphics[width=\linewidth]{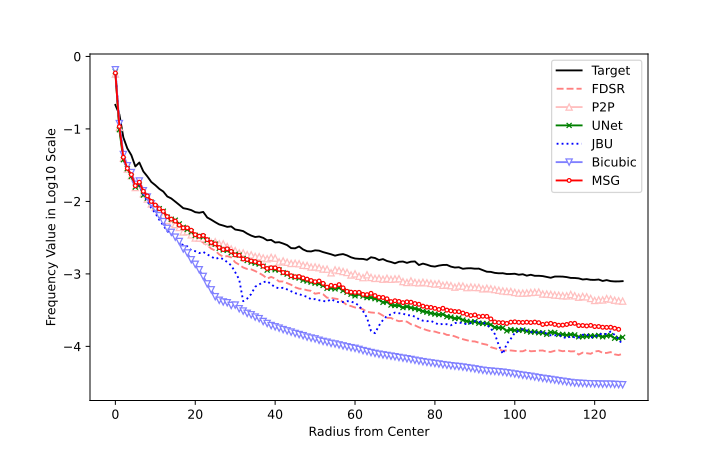}
    \caption{The frequency response of the outputs on the test sample in Figure \ref{fig:PredictionsForFixedSamples}.}
    \label{fig:FrequencyResponse_FixedSample}
\end{figure}

We also plot the frequency responses of the predicted maps alongside the ground truth. The prediction and target images are first transformed into the frequency domain via 2D Fourier transform and then converted from Cartesian to polar coordinates.
Since the radius in the 2D frequency domain corresponds to the magnitude of frequency components, the graph in Figure \ref{fig:FrequencyResponse_FixedSample} shows the average absolute values along different radii plotted against the radius values, showing how much information at a certain frequency is preserved. As a note, the cardinal sine (\textit{sinc}) pattern is clearly observable in Figure \ref{fig:FrequencyResponse_FixedSample} of JBU, which results from the Fourier transform of the rectangular-shaped behavior of the JBU output, shown in Figure \ref{fig:PredictionsForFixedSamples}.

\begin{figure}[htb!]
    \centering
    \includegraphics[width=\linewidth]{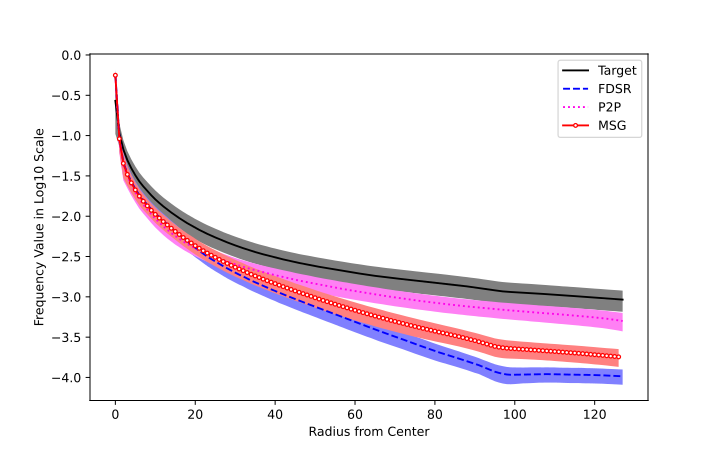}
    \caption{The histogram of the frequency responses of deep-learning-based guided super-resolution models.}
    \label{fig:FrequencyResponse_Hist}
\end{figure}

Figure \ref{fig:FrequencyResponse_Hist} is the histogram of the same information, depicting the standard deviation of the values across the different samples, calculated on the whole test set.
This graph also aligns with our earlier observation: MSG maintains HF components more effectively than FDSR. Also, P2P has more HF components than MSG. However, P2P’s accuracy is lower because it introduces new textures that are not present in the target map, leading to inconsistencies in the output. Lastly, we note that even though MSG shows a very convincing performance, Figure \ref{fig:FrequencyResponse_Hist} displays a large gap between the spectrum of the target biomass map and the prediction of MSG. This suggests that there is an avenue for further research, developing GSR methods specific to biomass estimation.

\subsection*{Our Findings}
\label{ssec:Findings}
\paragraph{Low-resolution maps as unexplored potential.}
The results presented in the previous section have successfully demonstrated that addressing biomass estimation as a GSR problem leads to significantly better performance. Conversely to direct regression from HR satellite observations, blending LR biomass information leads to roughly a $20\%$ reduction in RMSE compared to direct regression baselines (U-Net and ResNeXt). We argue that our proof-of-concept study on the BioMassters dataset encourages the exploration of a new paradigm for biomass estimation: to leverage low-resolution but high-quality information from global products like the ESA CCI AGB map to achieve very large-scale, accurate biomass maps at high spatial resolution.

\paragraph{Texture-copying is beneficial for GSR4B.}
Another important finding is that models, such as MSG, that do not specifically mitigate texture-copying actually perform better on our task than more advanced methods designed to suppress it. This difference likely stems from the distinct characteristics of our biomass data of the natural environment that contains much HF detail as compared to RGB$+$Depth data that often covers human-made objects and scenes with many planar surfaces. This suggests that there is potential for designing new approaches tailored specifically for super-resolving biomass data, other vegetation structure data with similar properties as well as more broadly environmental and climate data.

\subsection{Promising Future Research Directions}
\label{ssec:Extensions}

\paragraph{GSR for more applications in ecology and beyond.}
Our study indicates that the general idea of GSR applied for estimating ecological indicators at very high spatial resolution is promising. Learned spatial upsampling of existing large-scale products of lower spatial resolution using HR remote sensing data that carries HF information about the ecological indicator of interest provides a computationally lightweight strategy that can be applied to many forest structure variables and ecological indicators. Another line of research that seems worth investigating is whether a GSR-based method could update the LR source map if detailed changes (e.g., forest that is cut down) become visible in the HR satellite data. 

\paragraph{Generating an HR global ESA CCI AGB map.}
Our findings pave the way for exploring how GSR can help producing a global ESA CCI AGB map at higher spatial resolution. Using the ESA CCI AGB map as a source and combining it with HR satellite imagery as a guide and geo-referenced biomass datasets of HR seems a promising direction. 
A limiting factor in practice is, however, that such HR biomass datasets are not publicly available in large enough quantity and global distribution to enable proper validation of such an approach.
The larger domain gap between the large-scale LR source and more local HR target biomass map will also need further attention.   

\paragraph{Multi-temporal guidance.}
Another potential extension is incorporating time series data from satellite images. 
Here, we only used mono-temporal guide data, even for the U-TAE~\citep{Vivien2021UTAE} that would by construction allow time-series modeling.
Seasonal variations in forests and their differences across the globe can hardly be captured with a mono-temporal approach though but can carry additional information on the ecological variables of interest. Introducing satellite image time-series as a guide for upsampling biomass maps may therefore be an interesting future avenue of research.

In this paper, we have applied four different guided super-resolution methods, that were originally developed for RGB$+$Depth data in a computer vision context, to spatial upsampling of AGB with HR satellite imagery as a guide. Comparisons to more upsampling strategies and a detailed analysis indicate that learned GSR provides a promising direction toward HR yet global AGB maps based on existing, well-established products.  
Beyond this proof-of-concept for the specific case of AGB  map upsampling, learned GSR can possibly be applied similarly to more forest structure variables, ecological indicators, and possibly even climate data.  

\paragraph*{Acknowledgements.} 
This research benefited from funding from the EU Horizon Europe program under Grant Agreement No. 101131841. 
Additional funding for this project has been provided by the Swiss State Secretariat for Education, Research and Innovation (SERI).

{
\begin{spacing}{1.17}
    \normalsize
    \bibliography{biblio} 
\end{spacing}
}

\end{document}